\documentclass[sigconf]{acmart}
\usepackage{microtype}
\usepackage{hyperref}
\usepackage{float}
\usepackage{subcaption}
\usepackage{graphicx}  
\usepackage{todonotes}
\usepackage{enumitem}
\usepackage{comment}
\usepackage{times}
\usepackage{latexsym}
\usepackage{multirow}
\usepackage{tablefootnote}
\AtBeginDocument{%
  \providecommand\BibTeX{{%
    \normalfont B\kern-0.5em{\scshape i\kern-0.25em b}\kern-0.8em\TeX}}}

\copyrightyear{2022} 
\acmYear{2022} 
\setcopyright{acmcopyright}\acmConference[SIGIR '22]{Proceedings of the 45th International ACM SIGIR Conference on Research and Development in Information Retrieval}{July 11--15, 2022}{Madrid, Spain}
\acmBooktitle{Proceedings of the 45th International ACM SIGIR Conference on Research and Development in Information Retrieval (SIGIR '22), July 11--15, 2022, Madrid, Spain}
\acmPrice{15.00}
\acmDOI{10.1145/3477495.3531878}
\acmISBN{978-1-4503-8732-3/22/07}

\usepackage{xparse}
\NewDocumentCommand{\heng}
{ mO{} }{\textcolor{red}{\textsuperscript{\textit{Heng}}\textsf{\textbf{\small[#1]}}}}
\NewDocumentCommand{\avi}
{ mO{} }{\textcolor{blue}{\textsuperscript{\textit{Avi}}\textsf{\textbf{\small[#1]}}}}
\NewDocumentCommand{\arafat}
{ mO{} }{\textcolor{brown}{\textsuperscript{\textit{Arafat}}\textsf{\textbf{\small[#1]}}}}

\begin{document}
\fancyhead{}
\title{Entity-Conditioned Question Generation for Robust Attention Distribution in Neural Information Retrieval}

\author{Revanth Reddy}
\email{revanth3@illinois.edu}
\affiliation{
\institution{UIUC}
\country{United States}
}
\author{Md Arafat Sultan}
\email{arafat.sultan@ibm.com}
\affiliation{\institution{IBM Research AI}
\country{United States}
}
\author{Martin Franz}
\email{franzm@us.ibm.com}
\affiliation{\institution{IBM Research AI}
\country{United States}
}
\author{Avirup Sil}
\email{avi@us.ibm.com}
\affiliation{\institution{IBM Research AI}
\country{United States}
}
\author{Heng Ji}
\email{hengji@illinois.edu}
\affiliation{
\institution{UIUC}
\country{United States}
}


\renewcommand{\shortauthors}{Reddy, et al.}

\begin{abstract}
  We show that supervised neural information retrieval (IR) models are prone to learning sparse attention patterns over passage tokens, which can result in key phrases including named entities receiving low attention weights, eventually leading to model under-performance.
Using a novel targeted synthetic data generation method that identifies poorly attended entities and conditions the generation episodes on those, we teach neural IR to attend more uniformly and robustly to all entities in a given passage.
On two public IR benchmarks, we empirically show that the proposed method\footnote{Code and model available at: \href{https://github.com/blender-nlp/EntityConditionedQGen}{https://github.com/blender-nlp/EntityConditionedQGen}}
helps improve both the model's attention patterns and retrieval performance, including in zero-shot settings.
\end{abstract}




\maketitle



\section{Introduction}
\label{sec:intro}

Neural information retrieval (IR) performs query-passage matching at a semantic level, often using a dual-encoder architecture 
that encodes the queries and the passages separately.
Examples of such models include the Dense Passage Retriever (DPR) \cite{karpukhin2020dense} and ANCE \cite{xiong2020approximate}, which fine-tune transformer-based \cite{vaswani2017attention} pre-trained language models \cite{devlin2019bert} to compute contextualized representations of queries and passages.

\begin{figure}[t]
\begin{center}
   \includegraphics[width=1.0\linewidth]{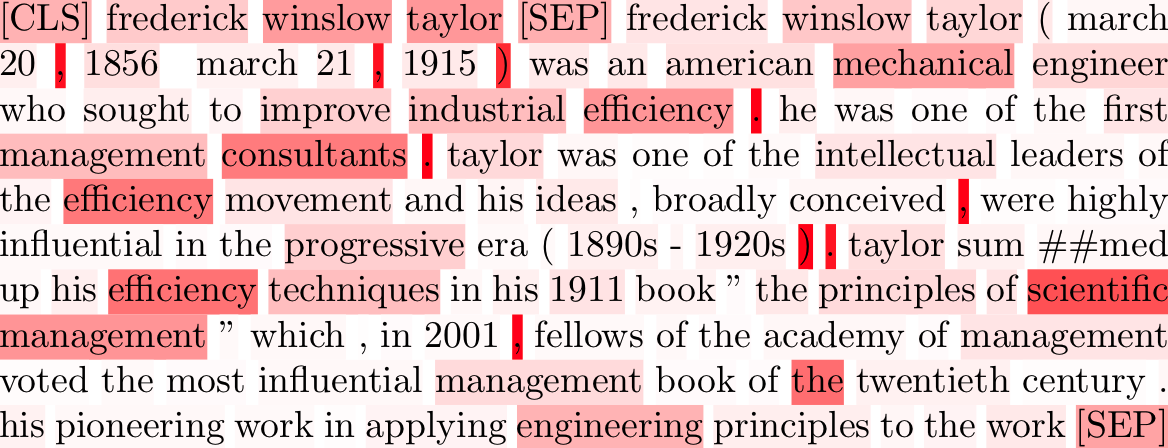}
\end{center}
   \caption{Heatmap of attention given to each token in DPR's passage representation. Darker shading indicates more attention.}
\label{fig:poorattn}
\end{figure}

In this paper, we first uncover a shortcoming in the passage encoder of such a dual-encoder IR model, namely DPR, which stems from its sparse attention pattern.
To illustrate, in Figure \ref{fig:poorattn} we show a heatmap of the attention weights of DPR's passage encoder over different tokens of an example passage (taken from the Natural Questions (NQ) dataset \cite{kwiatkowski2019natural}).
We can see that the attention given to many potentially important words and phrases, e.g, \textit{academy of management} and \textit{twentieth century}, are rather low.


\begin{table}[t]
    \centering
    \begin{tabular}{p{15.6em}|c|c}
    \multicolumn{1}{c|}{\textbf{Question}} & \multicolumn{1}{c|}{\textbf{Type}} & \textbf{Score}\\
    \hline
    \hline
    the \textcolor{blue}{\textit{american mechanical engineer}} who sought to improve \textcolor{blue}{\textit{industrial efficiency}} & \multirow{2}{*}{Gold} & \multirow{2}{*}{85.9} \\
    \hline
    who wrote the \textcolor{blue}{\textit{most influential management book}} of the \textcolor{blue}{\textit{twentieth century}} & \multirow{2}{*}{Synthetic} & \multirow{2}{*}{78.0} \\
    \hline
     who was considered the father of management during the \textcolor{blue}{\textit{progressive} era}    & \multirow{2}{*}{Synthetic} & \multirow{2}{*}{82.2} \\
     \hline
     who wrote the \textcolor{blue}{\textit{principles of scientific management}} & \multirow{2}{*}{Synthetic} & \multirow{2}{*}{86.8} \\
     \hline
    \end{tabular}
    \caption{Retrieval scores from DPR for the passage in Figure \ref{fig:poorattn}, against both a gold-standard question from NQ and three synthetic questions. The important terms in the question, that are also in the passage, are shown in \textcolor{blue}{\textit{italic}}.}
    \label{tab:analysis_questions}
\end{table}

\begin{figure*}[!htb]
\begin{center}
   \includegraphics[width=0.93\linewidth]{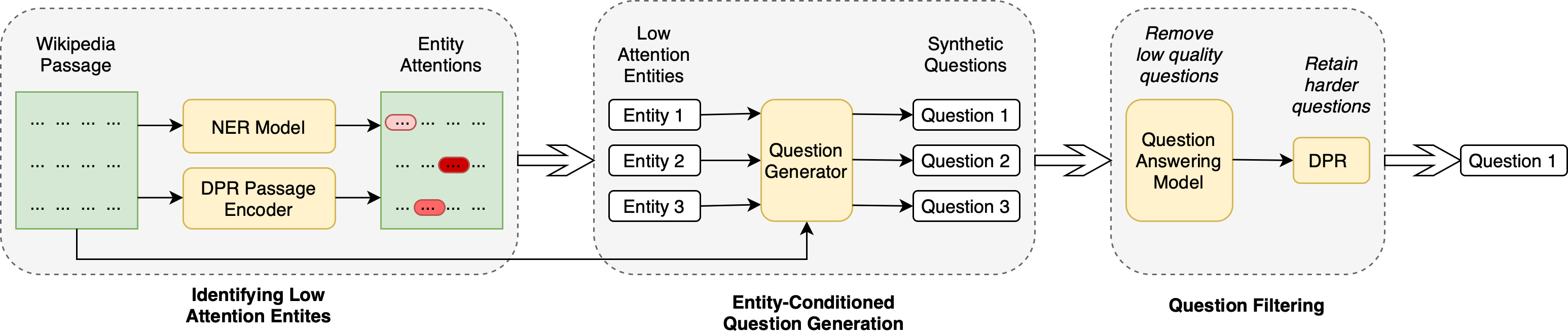}
\end{center}
   \caption{Overall framework of our synthetic data generation process to generate questions about named entities that receive low attentions from the DPR model.}
\label{fig:framework}
\end{figure*}

What is the effect of such attention, or lack thereof, on retrieval performance?
Table \ref{tab:analysis_questions} shows DPR's retrieval scores for a gold-standard question (from the NQ dataset) and three automatically generated synthetic questions (details in Section \ref{sec:method}) when paired with the passage of Figure~\ref{fig:poorattn}.
The gold-standard question, which overlaps highly with the well-attended first sentence of the passage, receives a relatively high retrieval score. 
Among the synthetic questions, the one that refers to the highest-attended entity (\textit{principles of scientific management}) gets the highest score, whereas the ones about less attended entities (\textit{twentieth century,  progressive era}) receive considerably lower scores.

To further quantify this, we randomly sampled 20k passages from Wikipedia and identified named entities that received the highest and lowest attentions from the DPR passage encoder (using the process described in Section \ref{sec:low_ent_attn}). We then generated synthetic questions corresponding to those entities (using the process of Section \ref{sec:cond_ques_gen}). We observe that on an average, the DPR score for questions corresponding to the highest attended entities was greater than that for questions corresponding to the lowest attended entities (73.7 vs. 72.1) in this sample. Further, we see the following pattern in the distribution of these entities: in a majority of cases (65\%), the highest attended entity in a given passage is present in the first half of the passage, whereas the lowest attended entity can be found more often in the second half of the passage (60\% of the cases).
These observations are indicative of certain biases present in DPR's passage encoder that prevent it from attending uniformly over the different named entities in an input passage.


As models trained on limited amounts of human-labeled data are prone to biases such as these, here we propose to augment the training data for neural IR with synthetic questions that are conditioned on the sparsely-attended parts of the passage.
Concretely, we generate questions specifically about entities that receive low attentions from the passage encoder of the neural IR model.
Our experiments show that augmenting the training with such questions does indeed enable neural IR models to attend more uniformly over passage tokens, resulting in performance improvements on multiple benchmark datasets.

Recently, there has been interest in understanding the robustness of neural IR models \cite{wu2021neural} and analyzing their behavior \cite{macavaney2022abnirml}. Our approach follows this line of work by leveraging the attentions of an IR model over given passages as a signal for better synthetic data augmentation. Prior work has also explored synthetic question generation for both question answering \cite{alberti2019synthetic, sultan2020importance, shakeri2020end} and neural information retrieval \cite{ma2021zero, gangi2021synthetic, reddy2021towards}; different approaches to generating questions from passages include: (a)~unconditioned generation \cite{gangi2021synthetic, reddy2021towards}, (b)~generation conditioned on the candidate answer phrases within the passage \cite{shakeri2020end, alberti2019synthetic}, and (c)~conditioned on the summary of the passage \cite{lyu2021improving, zhou2021generating}.
In contrast, our approach generates questions that are targeted towards the deficiencies of a given neural IR model, by conditioning the generation on sparsely attended entities in the passage. As we explain later in Section~\ref{sec:method}, our proposed methods for both identifying low-attention entities and generating questions that correspond to them can be leveraged for any dual-encoder IR model, including DPR \cite{karpukhin2020dense}, ANCE \cite{xiong2020approximate} or TAS-B \cite{hofstatter2021efficiently}. 


The main contributions of this paper are as follows:
\begin{itemize}
    \item We show that a SOTA neural IR model 
    is prone to learning sparse attention patterns over input passage tokens where key phrases (such as named entities) can receive low attention, leading to poor retrieval performance.

    \item We propose an entity-conditioned data augmentation strategy that generates questions about less attended entities in the passage.
    
    \item We demonstrate that incorporating these conditionally generated questions into the synthetic pre-training helps improve both model attention patterns and retrieval performance, including  zero-shot settings.
    

\end{itemize}

\section{Method}
\label{sec:method}

To help neural retrievers capture all entities in the passage, we propose to augment the training data with synthetic questions that are conditioned on the less attended entities in the passage. Our synthetic data generation process involves the following steps: (a) Identifying entities with low attention, (b) Generating questions that are conditioned on these entities, and (c) Filtering out low-quality synthetic questions. We describe each step in detail. 

\begin{figure*}[!htb]
\begin{center}
   \includegraphics[width=1.0\linewidth]{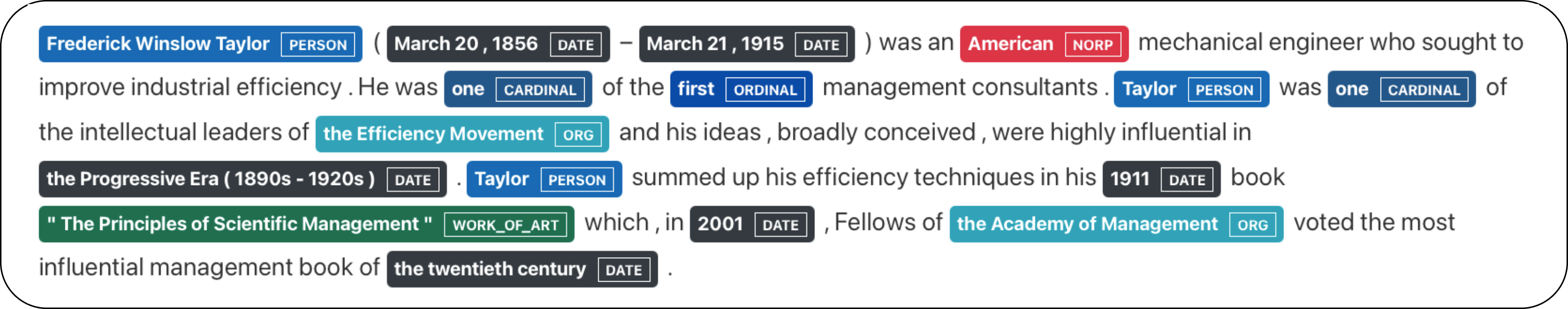}
\end{center}
   \caption{Entities automatically extracted from the passage of Figure \ref{fig:poorattn}.}
\label{fig:ieentities}
\end{figure*}

\subsection{Identifying entities with low attention}
\label{sec:low_ent_attn}
We use a named entity recognition system to first identify all the entities in a given passage (see Figure~\ref{fig:ieentities}). Then we compute attentions of the neural IR model over the passage and aggregate the attentions over the corresponding word-pieces to get the attention for each of the entities in the passage.
Finally, we identify the entities with the lowest attentions. Since DPR's passage encoder returns the CLS representation of the final layer as output, 
we use the attentions from that CLS token (i.e., where it serves as the query) as our passage attentions.

\begin{table}[!h]
    \centering
    \begin{tabular}{p{15.0em}|p{9.5em}}
    \multicolumn{1}{c|}{\textbf{Question}} & \multicolumn{1}{c}{\textbf{Conditioned Entity}} \\
    \hline
    \hline
     who was considered the father of management during the progressive era    & \multirow{2}{*}{Progressive Era} \\
     \hline
     who wrote the principles of scientific management & Principles of Scientific Management \\
     \hline
     who is known as the father of efficiency movement & \multirow{2}{*}{Efficiency Movement} \\
     \hline
    \end{tabular}
    \caption{Questions output by the synthetic generation system for the passage in Figure \ref{fig:ieentities}, based on the entity used for conditioning.}
    \label{tab:syn_questions}
\end{table}

\subsection{Entity-conditioned question generation}
\label{sec:cond_ques_gen}
Given a passage and an entity in that passage, we aim to generate a synthetic question about that entity using the passage. Specifically, we train a synthetic example generator to take a passage $p$, an entity $e$ and generate a question $q$ and its corresponding answer $a$. 
To achieve this, we fine-tune an encoder-decoder language model \cite{lewis2020bart} using examples from existing machine reading comprehension (MRC) datasets, which take the form of $(q,p,a)$ triples. 
Given such a triple, we first identify entities in $q$ that also appear in $p$. One such entity $e$ is passed as input along with $p$ to condition the question generation.
Following \citet{sultan2020importance}, we promote diversity in the generated questions by using top-$p$ top-$k$ sampling \cite{Holtzman2020The} during generation. Table \ref{tab:syn_questions} shows some generated questions conditioned on entities in the passage of Figure \ref{fig:ieentities}.

\subsection{Question filtering}
\label{question_filtering}
We employ a two-stage filtering process to promote high quality in the synthetic data. 
In the first stage, a generated question $q$ is considered to be consistent with the input passage $p$ if a separately trained MRC model can find an answer to $q$ in $p$ with high confidence.
All other questions are filtered out.
Among the remaining questions and their corresponding passages, we expect those to provide the best complementary signal (relative to existing gold-standard data) for which the baseline neural IR model has a low retrieval score. 
Hence, we only include such low scoring (harder) pairs in the synthetic pre-training set.

\section{Experimental Setup}
\label{sec:experiments}

\subsection{Datasets} 
We use the 21M Wikipedia passages from \citet{karpukhin2020dense} as the retrieval corpus for all our experiments. These passages come from the December 2018 Wikipedia dump, with each article split into text blocks of 100 words as passages, serving as our basic retrieval units. We use two public IR datasets in our experiments. \\

\noindent \textbf{\textit{Natural Questions:}} We train all systems on Natural Questions (NQ) \cite{kwiatkowski2019natural}, a dataset
with questions derived from Google's search log and  their human-annotated answers coming from Wikipedia articles. \citet{lewis2020question} report that 30\% of the NQ test set questions have near-duplicate paraphrases in the training set and 60--70\% of the test answers are also present in the training set. For this reason, in addition to the original 3,610 test questions, we also report evaluation on the non-overlapping subsets (1,313 no-answer overlap and 672 no-question overlap) released by \citet{lewis2020question}. \\


\noindent \textbf{\textit{WebQuestions:}} The dataset consists of questions obtained using the Google Suggest API, with answers selected from entities in Freebase by AMT workers \cite{berant-etal-2013-semantic}. We use the 2,032 test questions in this dataset for zero-shot evaluation.

\subsection{Baselines}
As traditional term matching baselines, we evaluate the TF-IDF system\footnote{\href{https://github.com/efficientqa/retrieval-based-baselines\#tfidf-retrieval}{https://github.com/efficientqa/retrieval-based-baselines\#tfidf-retrieval}} from \citet{chen2017reading} and the BM25 implementation provided by Pyserini\footnote{\href{https://github.com/castorini/pyserini/blob/master/docs/experiments-dpr.md}{https://github.com/castorini/pyserini/blob/master/docs/experiments-dpr.md}}. 
We evaluate DPR\footnote{\href{https://github.com/facebookresearch/DPR}{https://github.com/facebookresearch/DPR}} as our neural IR baseline\footnote{We note that our approach can be similarly applied to other dual-encoder IR models such as ANCE \cite{xiong2020approximate}.}.
\citet{karpukhin2020dense} report that the performance of DPR is affected by the number of in-batch negatives used in training, which in turn is dependent on the number of GPUs available.  
They use 128 in-batch negatives with eight 32GB V100s. Since we only had access to four 32GB V100s, we use 64 in-batch negatives. 
We call this implementation \textit{DPR (ours)}.

To compare our approach with a generation strategy that does not use any conditioning, we also train an unconditioned generation system, similar to \citet{reddy2021towards}, that generates question-answer pairs using just the passage as input. We call this the \textit{unconditioned} question generator, since the questions are not conditioned to be about any specific entities. This serves as a baseline question generation approach and is comparable with prior work \cite{ma2021zero, gangi2021synthetic, reddy2021towards} in synthetic data generation for IR, which do not enforce such specific conditioning into the question generation process.

\begin{table*}[!htb]
\centering
\begin{tabular}{l||cc|cc|cc||cc}
\multicolumn{1}{c||}{\multirow{2}{*}{Model}} & \multicolumn{6}{c||}{Natural Questions (NQ)}  & \multicolumn{2}{c}{WebQuestions} \\
\cline{2-9}
 & \multicolumn{2}{c|}{Full test} & \multicolumn{2}{c|}{No ans. overlap} & \multicolumn{2}{c||}{No ques. overlap} & \multicolumn{2}{c}{Test} \\ 
\hline
& Top-1 & Top-5  & Top-1 & Top-5 & Top-1 & Top-5 & Top-1& Top-5\\ 

TF-IDF & 14.2 & 32.0 & 13.6 & 28.6 & 14.6 & 31.8 & 14.5 & 32.1  \\
BM25 & 22.7 & 44.6 & 20.1 & 39.6 & 24.0 & 43.4 & 18.9 & 41.8 \\
DPR (ours)& 44.3 & 67.1 & 32.2 & 53.2 & 37.2 & 60.1 & 29.4 & 51.6 \\
UnCon-DPR & 45.8 & 68.4 & 32.7 & 54.4 & 36.9 & 60.6 & 31.5 & 53.2 \\
\hline
Mixed-DPR & \textbf{45.9} & \textbf{69.0} & \textbf{33.8} & \textbf{55.7} & \textbf{37.9} & \textbf{62.0} & \textbf{32.2} & \textbf{53.9} \\
\hline
\end{tabular}
\caption{Top-$k$ retrieval results (in \%) on test sets of Natural Questions (including the non-overlapping subsets of \citet{lewis2020question}) and WebQuestions. Numbers on WebQuestions are in zero-shot settings, since models have been trained on NQ.} 
\label{tab:final_num}
\end{table*}

\subsection{Synthetic Data Generation}
To create our synthetic pre-training corpus, first we derive a random sample of passages from the retrieval collection.
We identify the named entities in these passages using a publicly available NER system\footnote{\href{https://demo.deeppavlov.ai}{https://demo.deeppavlov.ai}} trained on the OntoNotes corpus \cite{weischedel2011ontonotes}. When selecting the entities that are used for conditioning, the following entity types are considered: Person, NORP, Facility, Organization, GPE, Location, Product, Event, Work of art, Law and Language. The MRC model used in the first stage of question filtering is trained sequentially on SQuAD2.0~\cite{rajpurkar2018know} for 3 epochs with a learning rate of 3e-05 and on Natural Questions for 1 epoch with a learning rate of 2e-05. We train the \textit{unconditioned} question generator by fine-tuning BART~\cite{lewis2020bart} with the question-passage-answer triples present in NQ. The model is trained for 3 epochs with a learning rate of 3e-05.

We fine-tune a separate BART model for \textit{conditioned} question generation, which takes a passage-entity pair as input and generates an entity-conditioned question and its answer as output. We repurpose the NQ dataset for training a conditioned question generation system, by converting the question-passage-answer triples into (question, conditioned entity, passage, answer) quadruples.
To obtain the conditioning entities used in training, we identify entities from noun chunks (obtained using spaCy \cite{honnibal2020spacy}) in the question that also occur in the corresponding passage.

We use the unconditioned generation system to first generate 1M synthetic training examples.
We then use the conditioned generation system to obtain 500k examples after filtering, and mix them with 500k unconditioned examples to obtain our final dataset of size 1M, which we call \textit{mixed} synthetic data. Since the conditioned data contains questions primarily about less attended entities, this combination with unconditioned examples
helps maintain adequate diversity in the final mixed dataset.  We follow the same process as in \citet{karpukhin2020dense} and use term matching to sample hard negatives for the questions.


\subsection{Training}
The DPR baseline is trained on NQ for 40 epochs with a learning rate of 2e-05. We name the model pre-trained on the 1M unconditioned synthetic data as \textit{UnCon-DPR} and the one pre-trained on the 1M mixed synthetic data as \textit{Mixed-DPR}. Pre-training is run for 10 epochs with a learning rate of 1e-05, following which both models are fine-tuned on NQ for 40 epochs with a learning rate of 2e-05.  

\section{Results}

Similar to \citet{karpukhin2020dense}, we evaluate all systems using top-$k$ retrieval accuracy, which is the percentage of questions with at least one answer in the top $k$ retrieved passages. Table \ref{tab:final_num} shows the results for the term matching and neural models.

Firstly, we can see that the two DPR models with synthetic pre-training improve over the baseline DPR system.
Our Mixed-DPR model, which employs entity-conditioned synthetic questions for pre-training, consistently outperforms all other models including UnCon-DPR, which is pre-trained only on unconditioned questions.
Crucially, on NQ, we observe greater improvements with Mixed-DPR on the non-overlapping and thus harder subsets of NQ, which indicates that the robustness of DPR improves with our proposed data augmentation strategy. Further, we see improvements for Mixed-DPR in a zero-shot evaluation on WebQuestions.

\subsection{Analysis}
To investigate the effect of the entity-conditioned questions used in synthetic pre-training, we examine how their application affects both the passage-level and token-level attentions of the DPR model.\\

\noindent \textbf{\textit{Passage-level attention distribution.}} First, we observe that the baseline DPR model (which is trained only on NQ) tends to attend more to the earlier sentences of a given passage.
We therefore compare attention on the first sentence (computed as the average attention over its tokens) with average attention on the rest of the sentences in the passage. We sample 10k passages from the retrieval corpus and compute attentions for the baseline DPR, UnCon-DPR and Mixed-DPR models. 
We observe that Mixed-DPR pays 1.8\% higher attention to the later sentences of the passage compared to the baseline DPR model. 
When compared to UnCon-DPR, this difference is 1.1\%. 
These results show that Mixed-DPR learns to attend more to the latter sentences of the passage which, as shown in Figure \ref{fig:poorattn}, is typically where most of the weakly attended entities of the baseline model occur. \\

\noindent \textbf{\textit{Token-level attentions.}} Here, we look at the entropy of token-level attentions in a given passage for the above models. Entropy here is a measure of the uniformity of a model's attention over the tokens in the passage, with a higher entropy indicating a more uniform distribution. For the 10k passages previously sampled, we see that the baseline DPR, UnCon-DPR and Mixed-DPR models have attention entropies of 3.97, 3.80 and 4.10 respectively, with Mixed-DPR being the highest. This suggests that the improvements in top-$k$ retrieval accuracy stem (at least partly) from a more scattered and potentially more robust attention pattern learned by Mixed-DPR.



\section{Conclusion}
We discover a specific issue in neural IR systems that stem from sparse attention patterns learned over input passage tokens, which can lead to sub-optimal performance on queries about less attended areas of the passage.
With targeted synthetic data augmentation, we address this issue for DPR---a state-of-the-art neural IR model---and enable it to attend more uniformly over passage tokens.
Our proposed method improves performance on two different datasets, and in in-domain as well as zero-shot evaluation.
While our work is an important first step towards solving this problem, one of the primary goals of this paper is to draw attention of the community to this important limitation of supervised neural IR and inspire future research on the topic. 
One potential direction is to incorporate additional objectives, e.g. multitask learning, to help models learn more robust attention patterns without requiring synthetic data. For example, named entity recognition as an auxiliary task may help the model identify key phrases in the passages, which in principle can help it to pay more attention to those during encoding.

\section{Acknowledgments}

We thank the anonymous reviewers for their valuable feedback and comments. This research is based upon work supported by U.S. DARPA AIDA Program No. FA8750-18-2-0014, U.S. DARPA KAIROS Program No. FA8750-19-2-1004. The views and conclusions contained herein are those of the authors and should not be interpreted as necessarily representing the official policies, either expressed or implied, of DARPA, or the U.S. Government. The U.S. Government is authorized to reproduce and distribute reprints for governmental purposes notwithstanding any copyright annotation therein.

\bibliographystyle{ACM-Reference-Format}
\bibliography{references}

\end{document}